\documentclass[letterpaper]{article} 
\usepackage[hyphens]{url}  
\usepackage{graphicx} 
\usepackage{graphicx}  
\usepackage{booktabs}
\usepackage{amsfonts}

\title{Multimodal Multitask Representation Learning \\ for Pathology Biobank Metadata Prediction}
\author{
Wei-Hung Weng\textsuperscript{\rm 1}\thanks{Work done at Google.}, 
Yuannan Cai\textsuperscript{\rm 2}, 
Angela Lin\textsuperscript{\rm 2}, \\
Fraser Tan\textsuperscript{\rm 2}, 
Po-Hsuan Cameron Chen\textsuperscript{\rm 2} \\
\textsuperscript{\rm 1}MIT CSAIL, Cambridge, MA, USA \\
\textsuperscript{\rm 2}Google Health, Mountain View, CA, USA \\
\texttt{ckbjimmy@mit.edu,} \\ \texttt{\{yuannan,linangela,frasert,cameronchen\}@google.com}
}

\begin{document}

\maketitle

\begin{abstract}
Metadata are general characteristics of the data in a well-curated and condensed format, and have been proven to be useful for decision making, knowledge discovery, and also heterogeneous data organization of biobank. 
Among all data types in the biobank, pathology is the key component of the biobank and also serves as the gold standard of diagnosis.
To maximize the utility of biobank and allow the rapid progress of biomedical science, it is essential to organize the data with well-populated pathology metadata. However, manual annotation of such information is tedious and time-consuming. 
In the study, we develop a multimodal multitask learning framework to predict four major slide-level metadata of pathology images. The framework learns generalizable representations across tissue slides, pathology reports, and case-level structured data.
We demonstrate improved performance across all four tasks with the proposed method compared to a single modal single task baseline on two test sets, one external test set from a distinct data source (TCGA) and one internal held-out test set (TTH). 
In the test sets, the performance improvements on the averaged area under receiver operating characteristic curve across the four tasks are 16.48\% and 9.05\% on TCGA and TTH, respectively. 
Such pathology metadata prediction system may be adopted to mitigate the effort of expert annotation and ultimately accelerate the data-driven research by better utilization of the pathology biobank.
\end{abstract}

\noindent Metadata have been proven to be useful for decision making and knowledge discovery~\cite{yee2003faceted,stephens2004utilizing}.
They also provide compact and domain-specific data representations for machine learning in different knowledge domains~\cite{linkert2010metadata,malet1999model,johnson2015love}. 
In the biomedical domain, metadata are critical for both data archiving and other downstream data-driven applications~\cite{posch2016predicting,weng2017medical}. 
One of the prominent applications is the development of biobanks~\cite{coppola2019biobanking}.

A biobank is a repository that stores biological tissue samples for research usage, and the essence of developing a biobank is to curate an organized dataset with well-populated metadata. 
Biobanks have been playing a critical role in numerous scientific and clinical breakthrough. 
For example, the development of one of the effective breast cancer immunotherapies, Herceptin, has greatly relied on metadata of a well-organized biobank to curate a cohort for the initial validation \cite{coppola2019biobanking}. 
The Cancer Genome Atlas (TCGA) and UK Biobank are also two large-scale repositories that support enormous studies and clinical trials to accelerate biomedical research~\cite{sudlow2015uk}. 
Biobanks usually contain a large amount of data from different modalities in heterogeneous formats, such as genome sequences, pathology images, and clinical reports.
To utilize these datasets, having well-populated metadata is necessary.

Pathology has been recognized as the gold standard for diagnosis across different medical specialties~\cite{kumar2014robbins}. 
In a biobank, pathology metadata along with the long-term follow-up survival data are the most valuable for modeling disease progression and patient outcome. 
However, they are also often the least organized data. 
This is primarily due to the variation of data archiving process across institutes and the tedious manual process of biological tissues in the modern pathology workflow.
Although an ideal approach for collecting accurate metadata is to ask pathologists for annotation while signing out the cases, it is infeasible for the already developed biobanks that have abundant longitudinal survival data. 
These data are also arguably of greater value due to the longer follow-up periods.  
Therefore, to maximize the utilization of the biobank, extensive manual pathology metadata annotation is required ~\cite{guo2016digital}.
Through advances in machine learning, modern computational techniques have shown promising results in automated prediction tasks.
For pathology, researchers have shown the capability pathologists-level performance on various tasks, such as tumor detection~\cite{litjens2017survey,coudray2018classification}, survival outcome prediction~\cite{nagpal2019development}, and even augmentation of the workflow through real-time feedback~\cite{chen2019augmented}. 
However, most works are based on the image modality alone without considering data from other modalities, such as free text reports.
Those works mostly focus on a single diagnosis task on few tissue types, different from a problem like biobank data curation that requires a well-performing model across multiple tasks on a wide range of specimens.
The heterogeneous, limited and imbalanced data also makes the automated pathology metadata prediction even more challenging.
Such a problem is also a common issue for machine learning and its downstream applications~\cite{lin2017focal}.

\begin{figure}[!t]
\centering
\includegraphics[width=0.8\linewidth]{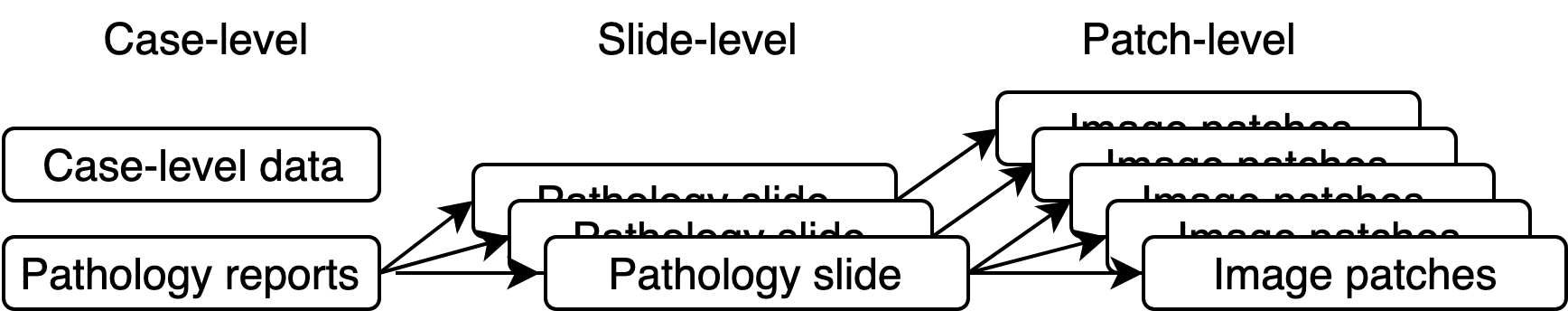} 
\hfill
\caption{Illustration of the information across the case, slide, and patch levels from a single patient.}
\label{fig1}
\end{figure}

In this study, we investigate multimodal multitask learning approach to jointly predict multiple slide-level metadata simultaneously from shared representation across image, text, and structured categorical variables in a limited and imbalanced sample size regime. 
Multitask learning leverages multiple prediction tasks to mitigate the issue of limited sample size since different tasks may share similar representations. 
Multimodal learning utilizes data of different modalities to learn a shared representation. 
Specifically, we incorporate case-level text from pathology reports with slide-level tissue images, because each of them holds different information that links to different metadata. 
Figure~\ref{fig1} and Appendix Table A1 are illustrations of the multi-level information in a single case.
We further conducted ablation analysis to investigate the importance and utility of different modalities. 
To this end, we make the following contributions in this study:
\begin{itemize}
    \item We propose a multimodal multitask learning framework using images, free texts and structured data for pathology metadata prediction with limited and imbalanced data.
    \item The proposed multimodal multitask framework outperforms the baseline single modal single task framework across all four pathology metadata prediction tasks.
    \item We observe the synergistic effect that by adding multimodal information on top of multitask framework outperforms either multimodal or multitask alone. 
\end{itemize}

\section{Related Works}
In this section, we start with a literature review on multitask learning and multimodal learning. 
Based on the existing works, we introduce challenges in multitask learning and multimodal learning, and justify the selected strategies for overcoming these challenges in the proposed framework.

\paragraph{Multitask Learning}
Multitask Learning (MTL), also known as joint learning or learning with auxiliary tasks, is a machine learning scenario that uses training signals from other related tasks to solve the harder tasks simultaneously~\cite{caruana1993multitask,ruder2017overview}.
MTL has been widely used in different domains such as natural language processing, speech, and computer vision~\cite{ruder2017overview}. 
It has also been applied to biomedical problems. 
One of the earliest applications is the pneumonia risk stratification task using the lab value prediction as the auxiliary MTL tasks~\cite{caruana1996using}. 
Despite the strong theoretical support for the utility of MTL, the performance gain of adopting MTL in the biomedical domain is not guaranteed~\cite{caruana1996using,nori2015simultaneous}. 

We argued that MTL can help leverage supervision signals from other tasks by training a model to predict correlated pathology metadata simultaneously from a shared representation. 
A natural question arises from this design is the ways to do parameter sharing. 
There are two main strategies for MTL to learn a shared representation, via hard parameter sharing or soft parameter sharing. 
Hard sharing strategy has a single pathway from input to a shared representation, and following the representation are task-specific heads with independent parameters. 
The approach enables lower layer parameters to be shared while parameters in each head to be task-specific. 
This enables learning a generalizable shared representation while also optimizing for the downstream tasks~\cite{caruana1993multitask}. 
In soft sharing strategy, each task has its pathway from input to output. 
The parameters for different pathways are soft-shared by imposing a joint regularization. 
This allows a certain degree of similarity across representations for different task pathways without enforcing them to be identical~\cite{duong2015low}.

Besides parameter sharing, another key aspect in MTL is the weighting of loss across different tasks. 
A prominent recent approach is a multi-objective optimization, which integrates the interactions between tasks into the loss fucntion~\cite{kendall2018multi}.
In this study, we incorporate the hard parameter sharing strategy with gradient-based multi-objective optimization to learn a better representation that can be shared across tasks. 
\begin{figure*}[!t]
\centering
\includegraphics[width=1.0\linewidth]{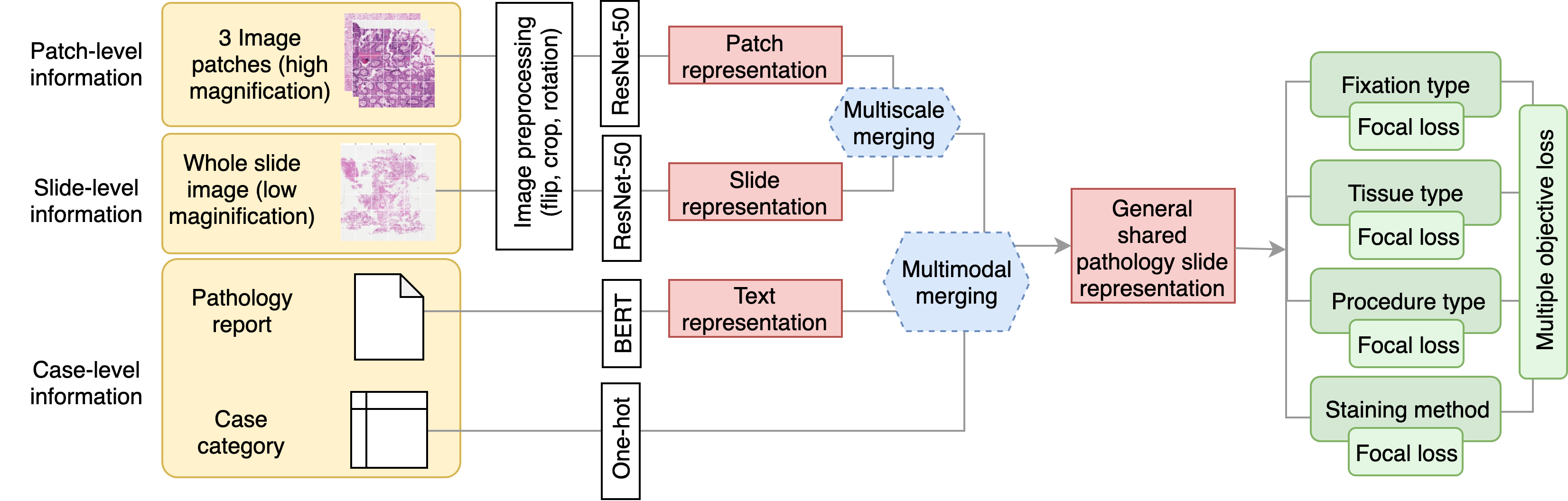} 
\hfill
\caption{Study overview and the proposed multimodal multitask learning framework. Color scheme:  input data (yellow), data encoder (white), multimodal merging operations (blue), representations (red), tasks (green), learning objectives (light green).}
\label{fig2}
\end{figure*}

\paragraph{Multimodal Learning}
The goal of multimodal learning is to learn a data representation capturing shared and independent information from data across different modalities.
There have been a wide success of multimodal learning in different fields~\cite{ngiam2011multimodal,baltruvsaitis2018multimodal}. 
In the biomedical domain, there is a strong need for such an approach because patient data usually come with multiple modalities such as waveforms, claims data, free texts, images, and genome sequences.
Researchers have utilized information from different modalities to approach various biomedical problems, such as tissue pattern recognition~\cite{schlegl2015predicting}, report generation~\cite{liu2019clinically}, medical language translation~\cite{weng2019unsupervised}, and clinical event prediction~\cite{suresh2017clinical}.
However, the key challenge of multimodal learning is the heterogeneity across different modalities due to vastly different statistical properties and varying levels of noise. 
In this study, we explore different strategies to effectively merge the heterogeneous modalities. 

There are two major approaches for multimodal learning: shared representation learning and cross-modal coordinated representation learning~\cite{ngiam2011multimodal}. 
Shared representation learning, i.e. learning a common embedding space, enforces the model to have a single latent representation from multimodal data. 
On the other hand, cross-modal coordinated representation learning, or embedding alignment, uses an additional step to align representations from different modalities with the assumption that the geometric structure of the representations is similar~\cite{chung2018unsupervised}. 

For shared representation learning, early fusion and late fusion are the two most prominent approaches~\cite{baltruvsaitis2018multimodal}.
Early fusion helps capture low-level interactions between modalities, which is ideal when there are dependencies between features in different modalities. 
Late fusion instead builds a meta-classifier to preserve more single-modal information since this approach doesn’t model the interactions between modalities at low-level. 

Since our goal is to learn a general representation from heterogeneous pathology images, texts, and structured data, we adopted the early fusion strategy to capture low-level interactions between modalities and keep the generalizability as much as possible.
We also investigated the model performance of using different operations for merging modalities.

\section{Methods}

\begin{figure}[htbp]
\centering
\includegraphics[width=0.9\linewidth]{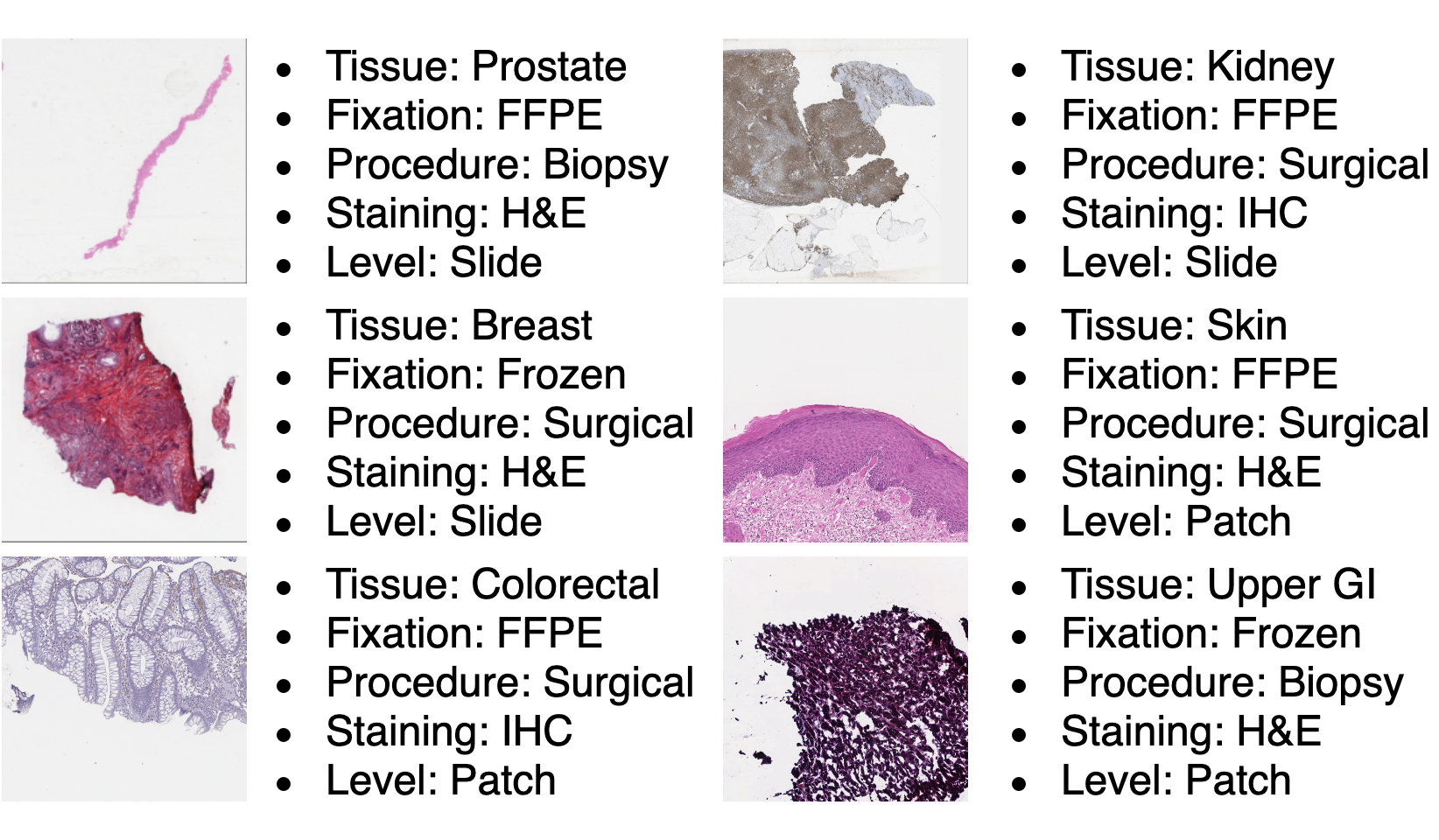} 
\hfill
\caption{Examples of pathology slide- and patch-level images with their metadata.}
\label{fig3}
\end{figure}

We propose a framework for joint prediction of pathology metadata while leveraging multimodal data, including images at different scales, texts from pathology reports, and the case-level structured data (Figure~\ref{fig2}).
We focus on four metadata commonly used for constructing research cohort from the pathology samples in biobanks. They are tissue type, fixation type, procedure type, and staining method of a slide. 
For example, to construct the research cohort for the lymph node metastasis detection, we need to identify samples of lymph node specimen (tissue type) using hemotoxylin and eosin staining (H\&E) (staining method) obtained by biopsy (procedure type) and fixed by formalin-fixed paraffin-embedded fixation (FFPE) (fixation type)~\cite{liu2017detecting}. 
Figure~\ref{fig3} shows samples of pathology images with their corresponding metadata. 
The proposed framework consisted of two main parts, a multitask output and a multimodal input. 
For the output, the model predicted four metadata tasks considering multiple objective loss. 
For the input, we adopted ResNet~\cite{he2016deep} and Bidirectional Encoder Representations from Transformers (BERT)~\cite{devlin2018bert} to extract features from slide images and free texts, respectively. 
For the case-level structured data such as primary cancer sites, we encoded them in the one-hot scheme. 
The intention of using multi-level data is to incorporate prior knowledge to the model for narrowing the prediction search space.
For example, using primary cancer sites as the input helps focusing the model on fewer tissue types that might occur in such case yet without the issue of data leakage from the pathology perspective (see Appendix A1). 

\paragraph{Multitask Learning}
We used MTL with hard parameter sharing strategy to learn the shared representation of pathology data from joint supervision across different tasks (Figure~\ref{fig3}). 
The proposed MTL loss function consists of two parts, a multi-objective loss, and a task-specific loss. 

%
For the multi-objective loss, to utilize the interactions between supervision from different tasks, we extended a Gaussian likelihood-based multitask loss accounting for all tasks simultaneously to mitigate the task-specific noise.
The approach models the estimated noise of task $t$ as a trainable parameter $\sigma_t$.
Different from treating all task equally without scaling the gradient of each task, the multi-objective loss has the potential to prevent overfitting to any specific task by considering the geometric average of all task-specific losses. This is different from the commonly seen arithmetic mean approach and addresses the issue of vastly different scales across task-specific losses. 

In more details, for the metadata prediction task $t$, we adopted a Gaussian likelihood to model the label $y_{it}$ and the predicted output from the neural network $f_t$ with training input $x_i$ and weights $w$. 
The output noise is modeled as a Gaussian distribution with zero mean and a standard deviation of $\sigma_t$. 
To find the stationary points of the Gaussian log likelihood function 
\begin{equation}
\sum_{it} \log(\frac{1}{\sqrt{2\pi} \sigma_t} \exp(-\frac{(y_{it} - f_t(x_i; w))^2}{2 \sigma_t^2}))
\end{equation}
, we set the partial derivation of the likelihood function with respect to $\sigma^2$ to zero, and obtain the update equation for $\sigma^2$ (see Appendix A2 for details).
By replacing $\sigma^2$ in the likelihood function, we obtain the loss function: 
\begin{equation}
\mathcal{L}_\mathrm{multi} = \sum_t \log ( \sum_i (y_{it} - f_t(x_i; w))^2)
\end{equation}
The loss function is a geometric average of the task-specific mean square error, different from the commonly-seen arithmetic mean. 

To mitigate the class imbalance issue, we applied the focal loss objective for the task-specific objectives~\cite{lin2017focal}.
In a binary classification task, we frequently use the cross entropy (CE) loss: 
\begin{equation}
\mathcal{L}_\mathrm{CE} = -\sum_i \log(p_i)
\end{equation}
, where $p_i=p$ if the label is correctly predicted, else $p_i=1-p$. $p \in [0, 1]$ is the estimated probability for correct class prediction.
The CE loss can be further extended to the $\alpha$-balanced CE loss that considers the class imbalance by multiplying a weighting factor $\alpha \in [0, 1]$: 
\begin{equation}
\mathcal{L}_{\alpha\mathrm{CE}} = -\sum_i \alpha_i \log(p_i)
\end{equation}
, where $a_i = a$ for correct prediction, else $a_i=1-a$.
However, both CE and $\alpha$-balanced CE loss can not differentiate easy and hard samples well. 
Thus focal loss is introduced to reshape the $\alpha$-balanced CE loss to down-weight easy samples and focus on hard samples by introducing a modulating factor $(1-p_i)^{\gamma}$ with a focusing parameter $\gamma \geq 0$.
The objective can be expressed as 
\begin{equation}
\mathcal{L}_\mathrm{focal} = - \sum_i \alpha_i (1-p_i)^{\gamma} \log(p_i)
\end{equation}
When $\gamma > 0$, the loss contribution of easy samples will be discounted, else the loss function will turn into CE loss.

The final loss function in our MTL framework is the combination of the task-specific losses and the multi-objective loss:
\begin{equation}
\mathcal{L} = \mathcal{L_{\mathrm{multi}}} + \sum_t \mathcal{L_{\mathrm{focal}}}_t,
\end{equation}
where $t$ is the task index.

\paragraph{Multimodal Learning}
To develop a joint representation across modalities, we explored two early fusion methods.
We investigated a widely-used vector concatenation and the compact bilinear pooling (CBP) that captures interactions between modalities more expressively~\cite{fukui2016multimodal}.

Vector concatenation is well-known as a strong approach to merge modalities. For the vector concatenation method, the shared representation used for the downstream tasks is derived as the following: $\mathcal{V}_{\mathrm{shared}} = [\mathcal{V}_{\mathrm{image}}; \mathcal{V}_{\mathrm{text}}; \mathcal{V}_{\mathrm{structured}}]$, where $\mathcal{V}$ is the vector representation of the modality. 

Bilinear models take the outer product of two representations to form the joint representation. It is an alternative approach to learning the interaction between two vector representations but with a huge computational cost.
~\cite{fukui2016multimodal} proposed CBP to mitigate the issue of the computational heavy outer product by adopting Fourier transformation tricks to operate in the transformed space. Besides, CBP also utilizes a Count Sketch projection (CSP) function $\Psi$ to project the high-dimension outer product vector to a lower-dimensional space as a more compact representation. 


In details, the inputs of CSP function are the outer product vector $v \in \mathbb{R}^n$ and two randomly uniformly initialized constant vectors $s \in \{-1, 1\}^n$ and $h \in \{1, ..., d\}^n$. 
The function outputs a latent representation vector $\mathcal{V}_{\mathrm{shared}} \in \mathbb{R}^d$, where $n\gg d$.
The $j$-th element of $\mathcal{V}_{\mathrm{shared}}$, $\mathcal{V}_{\mathrm{shared}}(j)$, is defined as 
\begin{equation}
\mathcal{V}_{\mathrm{shared}}(j) = \sum_{i\in \{i|h_i = j\}} s_i \times v_i
\end{equation}
%
A CSP function of an outer product between two vectors $\mathbf{X}, \mathbf{Y}$ is equivalent to the convolution between the CSP applied $\mathbf{X}$ and $\mathbf{Y}$. Precisely, $\mathcal{V}_{\mathrm{shared}} = \Psi(\mathbf{X} \otimes \mathbf{Y}, s, h) = \Psi(\mathbf{X}, s, h) * \Psi(\mathbf{Y}, s, h)$, where $\otimes$ is the outer product, $*$ is the convolution operation. This can again be rewritten in a format using Fourier transformation ($\mathrm{FFT}$) and inverse Fourier transformation ($\mathrm{FFT}^{-1}$), such that $i*j = \mathrm{FFT}^{-1}(\mathrm{FFT}(i) \odot \mathrm{FFT}(j))$, where $\odot$ is the element-wise product. 
Thus, the original $\Psi$ function for the dimensionality reduction of outer product can be operated using Fourier transformation as following:
\begin{equation}
\mathcal{V}_{\mathrm{shared}} = \mathrm{FFT}^{-1}(\mathrm{FFT}(\Psi(\mathbf{X}, s, h)) \odot \mathrm{FFT}(\Psi(\mathbf{Y}, s, h)))
\end{equation}
, where $\mathbf{X}, \mathbf{Y}$ are input vectors and $s, h$ are randomly assigned vectors mentioned above. These transformation tricks have the benefits of reduced computation and memory usage, enabling operations on high-dimensional vectors. 

\paragraph{Multiscale Imaging}
We also explored multimodal learning in the context of multiscale imaging.
This is inspired by pathologists' workflow in examining images at different image magnifications to get both the context and the details.
We used the whole slide images in low magnification and three high magnification image patches randomly cropped from the tissue area in the slide image.
We applied the vector concatenation and CBP for multiscale learning. 
Using the vector concatenation, the image representation $\mathcal{V}_{\mathrm{image}}$ will be:
\begin{equation}
\mathcal{V}_{\mathrm{image}} = [\mathcal{V}_{\mathrm{slide}}; \mathcal{V}_{\mathrm{patch}_1}; \mathcal{V}_{\mathrm{patch}_2}; \mathcal{V}_{\mathrm{patch}_3}]
\end{equation}
If using CBP, 
\begin{equation}
\mathcal{V}_{\mathrm{image}} = \mathrm{CBP}([\mathcal{V}_{\mathrm{slide}}, [\mathcal{V}_{\mathrm{patch}_1}; \mathcal{V}_{\mathrm{patch}_2}; \mathcal{V}_{\mathrm{patch}_3}])
\end{equation}
, where $\mathrm{CBP}(\cdot)$ is the CBP operation described above. 
For CBP, we followed the parameter setting in~\cite{fukui2016multimodal} with 16000-D output representation. 

\paragraph{Natural Language Representation}
To extract representation from the free text pathology reports, we used an attention-based BERT model as the encoder~\cite{devlin2018bert}.
The implementation of the BERT encoder contains 12 Transformer blocks and 12 self-attention heads with 768-D representation.
The encoder is pretrained on large English corpora consisting of Wikipedia and BookCorpus. 
The last two Transformer blocks were set to be trainable to be fine-tuned for our tasks. 
After integrating the pretrained BERT into the proposed multimodal merging and multitask framework, two Transformer blocks were fine-tuned with the backpropagated gradient from the overall loss (Figure~\ref{fig3}).

\section{Experiments}

\paragraph{Datasets}
We collected two datasets for this study, a dataset from a tertiary teaching hospital (TTH) and a publicly available TCGA dataset. 
The metadata of two datasets has been annotated by board-certified pathologists.
The TTH dataset was split into three subsets of 80\% (18,413 slides / 4,972 cases), 10\% (2,570 slides / 1,324 cases), 10\% (2,570 slides / 1,324 cases) for training, validation and testing, respectively. We adopted iterative stratified sampling to ensure that the proportions of classes are nearly equally distributed in three subsets.
A random subset of the TCGA dataset (10,440 slides / 7,084 cases) was annotated and used as an independent hold-out test set to evaluate the generalization of our method to the unseen data source. 

All samples were categorized into two fixation types, 14 tissue types (with one type as ``others''), two procedure types, and two staining methods as shown in Table~\ref{tab1}. 
Extremely rare tissue types 
are recategorized into ``others'' unless they are the top-8 tissue types in TCGA dataset. 

\begin{table}[]
\centering
\resizebox{\columnwidth}{!}{
\begin{tabular}{ccrrrrrrrr}
\toprule
Task & Dataset & \multicolumn{6}{c}{TTH} & \multicolumn{2}{c}{TCGA} \\
\midrule
 & Split & \multicolumn{2}{c}{Train} & \multicolumn{2}{c}{Val} & \multicolumn{2}{c}{Test} & \multicolumn{2}{c}{Test} \\
\midrule
 & Count/\% & N & \% & N & \% & N & \% & N & \% \\
\midrule
 & \#Case & 4972 &  & 1324 &  & 1432 &  & 7084 &  \\
 & \#Slide & 18413 &  & 2570 &  & 2835 &  & 10440 &  \\
\midrule
Fixation & FFPE & 17790 & 96.6 & 2409 & 93.7 & 2660 & 93.8 & 4734 & 45.3 \\
 & Frozen & 623 & 3.4 & 161 & 6.3 & 175 & 6.2 & 5706 & 54.7 \\
\midrule
Tissue & LN & 2561 & 13.9 & 324 & 12.6 & 394 & 13.9 & 0.0 & 0.0 \\
 & Uterus/cervix & 1697 & 9.2 & 269 & 10.5 & 263 & 9.3 & 701 & 6.7 \\
 & Breast & 2029 & 11.0 & 298 & 11.6 & 301 & 10.6 & 1010 & 9.7 \\
 & Other & 2131 & 11.6 & 242 & 9.4 & 264 & 9.3 & 3284 & 31.5 \\
 & Skin & 1966 & 10.7 & 235 & 9.1 & 290 & 10.2 & 178 & 1.7 \\
 & Prostate & 2048 & 11.1 & 279 & 10.9 & 305 & 10.8 & 465 & 4.5 \\
 & Colorectal & 1739 & 9.4 & 241 & 9.4 & 282 & 9.9 & 707 & 6.7 \\
 & H\&N & 1610 & 8.7 & 221 & 8.6 & 243 & 8.6 & 403 & 3.9 \\
 & Thyroid & 1025 & 5.6 & 161 & 6.3 & 151 & 5.3 & 447 & 4.3 \\
 & UGI & 960 & 5.2 & 147 & 5.7 & 174 & 6.1 & 589 & 5.6 \\
 & Ovary & 581 & 3.1 & 89 & 3.5 & 104 & 3.7 & 495 & 4.7 \\
 & Kidney & 29 & 0.2 & 29 & 1.1 & 29 & 1.0 & 1130 & 10.8 \\
 & Lung & 37 & 0.2 & 35 & 1.4 & 35 & 1.2 & 1031 & 9.9 \\
\midrule
Procedure & Surgical & 10853 & 58.9 & 1435 & 55.8 & 1651 & 58.2 & 9551 & 91.5 \\
 & Biopsy & 7560 & 41.1 & 1135 & 44.2 & 1184 & 41.7 & 889 & 8.5 \\
\midrule
Staining & H\&E & 15086 & 81.9 & 1787 & 69.5 & 1963 & 69.2 & 10425 & 99.9 \\
 & IHC & 3327 & 18.1 & 783 & 30.5 & 872 & 30.8 & 15 & 0.01 \\
\bottomrule
\end{tabular}
}
\caption{Datasets statistics and label distributions. Please refer to Appendix Table A2 for abbreviations.}
\label{tab1}
\end{table}

\paragraph{Data Preprocessing and Resampling}


In multitask learning, class imbalance issue is aggravated concerning the number of tasks. 
In this study, there are 112 possible label combinations across four tasks (2 fixation types $\times$ 14 tissue types $\times$ 2 procedure types $\times$ 2 staining methods), Table~\ref{tab1}.
The ratio of cases between the most frequent combination and the least frequent combination grows exponentially with respect to the number of tasks.
For example, the number of cases in the most common combination (FFPE, LN, Surgical, H\&E) is 16,415 times more than the least frequent combination (Frozen, Lung, Biopsy, IHC). 
To address this issue for model development, we upsampled rare combinations to 500 cases per combination and downsampled combinations with abundant data to 1000 cases per combination in the training set. 
There is no change to the class distribution for validation and test sets. 


We used images with two different scales in this study, low magnification whole slide images and high magnification image patches from slides.
For whole slide images, due to the difficulty of fitting gigapixel images into the memory~\cite{liu2017detecting}, all the images were retrieved at $0.3125\times$ and then rescaled to $512 \times 512$ pixels by bilinear interpolation regardless the aspect ratio. 
The slide-level features such as spatial relationship and colors of the tissues were still preserved through the rescaling.
From the pathology perspective, such preprocessing does not affect the decision of metadata annotation because the spatial relationship is preserved.
We used the high magnification image patches for incorporating fine-grained features. 
The images are retrieved from the pathology slide images at $5\times$ magnification and randomly cropped out three patches within the tissue area at $299 \times 299$ pixels.

For the free-text pathology reports, we set a fixed text sequence length of 64 for the BERT encoder input. 
The average sequence length of the pathology reports in our datasets is around 100, yet the report lengths are skewed since most biopsy reports are very short. 
Also, the most important information in the pathology reports is usually shown at the beginning, such as the diagnosis of the case. 
Thus, we set the maximal sequence length at around the 60\% quantile of all report lengths for text encoding. 
We also selected the primary cancer site as structured data input to capture more related information at the case level (see Appendix A1 for more details).
We used a one-hot representation to encode the case-level structured input.

\paragraph{Neural Network Details and Baseline}
We used the standard ResNet-50 model for learning the slide- and the patch-level image representations~\cite{he2016deep}.
Text representation is learned from the BERT encoder~\cite{devlin2018bert}.
Case-level structured data are featurized with one layer fully connected network.

We applied the focal loss with $\gamma=2.0, \alpha=0.5$, and used the Adam optimizer with $\beta_1=0.9, \beta_2=0.999, \epsilon=10^{-8}$ and set a clipping normalization value of 0.5 for optimization. 
Exponential decay learning rate scheduling with an initial learning rate of $10^{-3}$, with a decay rate of $0.9$ for every $200$ training steps, were used. 
We trained the model with five epochs with a batch size of 32. 

The baseline configuration for comparison is the single modal single task framework with multiscale image (slide + patch) inputs.
We used both slide and patch-level information as the baseline since it is the most comprehensive way to read the slide for pathologists.

\paragraph{Evaluation}
We evaluated the models' performance on metadata prediction with two standard metrics, the macro average of the area under ROC curve (AUC-ROC) and the macro average of the area under the precision-recall curve (AUC-PR). 
The AUC-ROC measures the probability of the model ranking a randomly chosen positive sample higher than a randomly chosen negative sample. AUC-PR measures the trade-off between precision (positive predictive value) and recall (sensitivity). 
Macro AUC-ROC and macro AUC-PR take an equally weighted average of AUC-ROCs and AUC-PRs across all classes, respectively. An equal-weighted average is desired because rare cases are treated equally as common cases. 
Other commonly used metrics such as accuracy, micro average AUC-ROC, and micro average AUC-PR are not suitable for this study due to the highly imbalanced test set. 
These metrics are computed per sample instead of per class, therefore, a model can achieve high accuracy simply by predicting well on the majority classes while ignoring minority classes.

\section{Results and Discussions}

\paragraph{Metadata Prediction with Multimodal Multitask Learning} 
We start with a comparison between the proposed multimodal multitask framework and the baseline single modal single task benchmark in Table~\ref{tab2}. 
The proposed method (MM) achieved higher AUC-ROCs compared to the baseline (SS) on three out of four tasks and on-par on one task, on the hold-out independent TCGA test set (Table~\ref{tab2}, $\Delta$(MM, SS) = 35\% (tissue), 2.33\% (fixation), 0\% (procedure), 1.15\% (staining)). 
On the TTH test set, MM achieves higher AUC-ROCs on two tasks, on-par on one task, and slightly lower on one task (Table~\ref{tab2}, $\Delta$(MM, SS) = 23.94\% (tissue), -1.01\% (fixation), 4.41\% (procedure), 0\% (staining)). 
Predicting tissue type is the task with the great improvement by using multimodal multitask framework. 

We found that the main contribution of the performance improvement comes from the multimodal component instead of the multitask component (Table~\ref{tab2}, $\Delta$(MS, SS) = 23.58\% (tissue), 1.40\% (fixation), -0.76\% (procedure), -2.92\% (staining) versus $\Delta$(SM, SS) = 2.82\% (tissue), -0.65\% (fixation), -1.47\% (procedure), 0.60\% (staining)). 
However, combining multitask with multimodality often lead to further increases in performance (Table~\ref{tab2}, $\Delta$(MM, MS) = 7.32\% (tissue), -0.60\% (fixation), 2.99\% (procedure), 3.60\% (staining)). 
It allows the multitask framework to utilize the inductive bias from one task to learn a more generalized representation that improves or keeps the performance of other tasks. 
Such effect is consistent with the argument that shared representation learned from multitask helps generalize across tasks, especially when the major information sources are slightly different but correlated between tasks.
In our case, image mainly provides the evidence for fixation and staining prediction, whereas texts are informative for tissue type prediction, incorporating both multimodal and multitask framework helps learn a better representation that share more underlying patterns in pathology.

Among the four prediction tasks, tissue type is the most challenging task for pathologists. 
To classify tissue type from images, a pathologist will need to review the tissue morphology in both low and high magnification while incorporating prior knowledge about the case from the pathology report. 
On the other hand, the other three tasks are relatively simple and can be learned by a layman with appropriate training.
Furthermore, tissue type prediction task has 14 classes while the other tasks are binary classification problems. 
Due to the intrinsic difficulty of the task, the baseline only reached an AUC-ROC of 0.60 on TCGA test set, while the multimodal multitask framework significantly improved the performance to an AUC-ROC of 0.81 (Table~\ref{tab2}, TCGA MM), and up to 0.92 after further optimization (Table~\ref{tab5}).

We observed that the majority of the performance gain came from the incorporation of multimodality instead of multitask. 
This is because that the key information includes in the reports are pathology findings, which are highly relevant to tissue type, and partially related to the procedure since different procedures may provide different findings mentioned in the reports. 
The reports and case-level information are less informative for fixation and staining prediction since these two metadata are directly related to slide images and less emphasized in the reports.
However, comparable results are still observed since MTL can utilize the tissue type prediction as an auxiliary task to keep or improve the model prediction power for other tasks.
Therefore, we argue that the additional modalities, free text, and structured data, provide useful information about the specific task of tissue type prediction. 
This is also intuitively reasonable because understanding the pathology reports provides strong prior for a model to predict tissue type among fewer options consistent with the report.
\begin{table}[]
\centering
\resizebox{\linewidth}{!}
{
\begin{tabular}{cccccc}
\toprule
Dataset & Cfg & Tissue & Fixation & Procedure & Staining \\
\midrule
TCGA & SS & 0.60 & 0.86 & 0.67 & 0.87 \\
(external) &    & (0.51, 0.67) & (0.84, 0.88) & (0.60, 0.72) &  (0.16, 1.00) \\
 & SM & 0.60 & 0.84  & 0.65 & 0.87  \\
 &    & (0.57, 0.62) & (0.81, 0.86) & (0.58, 0.71) & (0.18, 1.00) \\
 & MS & 0.79 & \textbf{0.91}  & 0.65 & 0.84  \\
 &    & (0.72, 0.85) & \textbf{(0.90, 0.93)} & (0.58, 0.71) & (0.22, 1.00) \\
 & MM & \textbf{0.81} & 0.88  & \textbf{0.67} & \textbf{0.88}  \\
 &    & \textbf{(0.79, 0.83)} & (0.85, 0.90) & \textbf{(0.61, 0.72)} & \textbf{(0.24, 1.00)} \\
 \midrule
 TTH & SS & 0.71 & 0.99  & 0.68 & 0.84  \\
 &    & (0.60, 0.81) & (0.98, 1.00) & (0.60, 0.75) & (0.77, 0.91) \\
 & SM & 0.75 & \textbf{1.00}  & 0.67 & \textbf{0.85}  \\
 &    & (0.69, 0.80) & \textbf{(0.98, 1.00)} & (0.58, 0.75) & \textbf{(0.78, 0.91)} \\
 & MS & 0.82 & 0.96  & 0.69 & 0.82  \\
 &    & (0.75, 0.89) & (0.87, 1.00) & (0.60, 0.77) & (0.75, 0.89) \\
 & MM & \textbf{0.88} & 0.98  & \textbf{0.71} & 0.84  \\
 &    & \textbf{(0.85, 0.91)} & (0.94, 1.00) & \textbf{(0.63, 0.79)} & (0.77, 0.90) \\
\bottomrule
\end{tabular}
}
\caption{Quantitative evaluation across different setups on the TCGA and TTH test set. TCGA is an independent test set with different data distribution from the TTH dataset. We reported the values of macro AUC-ROC with 95\% confidence intervals. For evaluations of macro AUC-PR on test and validation sets, please refer to the Appendix Section A3. Abbreviations: configurations (Cfg), single modal single task (SS), single modal multitask (SM),  multimodal single task (MS), multimodal multitask (MM).}
\label{tab2}
\end{table}

\paragraph{Utility of the Multitask Framework Alone}
Although the multimodal multitask learning approach showed improved performance over the baseline, limited improvements are seen if we consider multitask learning alone without multimodal inputs (Table~\ref{tab2}, $\Delta$(SM, SS) = 2.82\% (tissue), -0.65\% (fixation), -1.47\% (procedure), 0.60\% (staining), where the performance of the two approaches are on-par considering the confidence interval). 
This is likely due to insufficient information in the image modality alone for some tasks, such as predicting tissue type from the image alone, which is regarded as a nontrivial task by pathologists. 
Although multitask learning alone does not lead to a significant improvement over the baseline, it demonstrates on-par performance using only one model with four task heads instead of four independent models. 
This indicates that the multitask framework yields a more generalized representation for different tasks while greatly reduces the required computation resource for model development and speed up model iterations and inferences.

\paragraph{Comparison of Multimodality Strategies}
For the multimodality module, we also investigated different merging strategies of integrating image and text information under multitask scenario. 
Table~\ref{tab3} shows the performance comparison of vector concatenation and CBP. 
We observe higher performance with vector concatenation in most tasks except for the tissue type prediction problem. 
In~\cite{fukui2016multimodal}, CBP worked reasonably well for visual-text question answering problems which image and text modalities both have a good correlation on the targeted task. 
Similarly, in this study, for tissue type prediction which requires both image and text modalities, CBP shows an improved performance by leveraging both modalities. 
On the other hand, for tasks that a single modality is sufficient and other modalities are not expected to contain task-related information, CBP yield inferior performance.
For example, text reports usually do not contain information about fixation and staining (AUC-ROC of 0.59 using only text modality on TCGA). 
Therefore, we observed worse performance with CBP relative to the concatenation method (Table~\ref{tab3}). 
\begin{table}[]
\centering
\resizebox{\linewidth}{!}
{
\begin{tabular}{cccccc}
\toprule
Dataset & MM-Strategy & Tissue & Fixation & Procedure & Staining \\
\midrule
TCGA & Concat & 0.81 (0.79, 0.83) & \textbf{0.88 (0.85, 0.90)} & \textbf{0.67 (0.61, 0.72)} & \textbf{0.88 (0.24, 1.00)} \\
 (external)  & CBP & \textbf{0.86 (0.85, 0.87)} & 0.52 (0.48, 0.55) & 0.59 (0.53, 0.64) & 0.40 (0.06, 0.79) \\
\midrule
TTH & Concat & 0.88 (0.85, 0.91) & \textbf{0.98 (0.94, 1.00)} & \textbf{0.71 (0.63, 0.79)} & \textbf{0.84 (0.77, 0.90)} \\
 & CBP & \textbf{0.89 (0.85, 0.92)} & 0.52 (0.32, 0.71) & 0.50 (0.41, 0.60) & 0.46 (0.36, 0.55) \\
\bottomrule
\end{tabular}
}
\caption{Performance comparison between different representation merging methods on the testing set, TCGA and TTH. We reported the values of macro AUC-ROC with 95\% confidence intervals.}
\label{tab3}
\end{table}

\paragraph{Ablation Analysis to Understand the Importance of Different Modalities}
To explore the effect of each modality, we conducted ablation analysis by removing text and structured data one at a time while keeping the whole slide image modality (Table~\ref{tab4}). 
Whole slide image modality is not removed because it is the base component for the pathology metadata prediction.
\begin{table}[]
\centering
\resizebox{\linewidth}{!}
{
\begin{tabular}{cccccc}
\toprule
Dataset & Modality & Tissue & Fixation & Procedure & Staining \\
\midrule
TCGA & All & \textbf{0.81 (0.79, 0.83)} & \textbf{0.88 (0.85, 0.90)} & \textbf{0.67 (0.61, 0.72)} & \textbf{0.88 (0.24, 1.00)} \\
 (external) & All w/o text & 0.66 (0.64, 0.68) & 0.84 (0.82, 0.86) & 0.66 (0.60, 0.72) & 0.86 (0.17, 1.00) \\
 & All w/o structured & 0.76 (0.74, 0.78) & 0.85 (0.83, 0.88) & 0.60 (0.54, 0.66) & 0.87 (0.30, 1.00) \\
\midrule
TTH & All & \textbf{0.88 (0.85, 0.91)} & \textbf{0.98 (0.94, 1.00)} & \textbf{0.71 (0.63, 0.79)} & \textbf{0.84 (0.77, 0.90)} \\
 & All w/o text & 0.78 (0.74, 0.83) & 0.98 (0.94, 1.00) & 0.69 (0.60, 0.77) & 0.84 (0.77, 0.90) \\
 & All w/o structured & 0.85 (0.82, 0.89) & 0.98 (0.94, 1.00) & 0.70 (0.61, 0.77) & 0.83 (0.76, 0.90) \\
\bottomrule
\end{tabular}
}
\caption{Ablation analysis by removing input modality one at a time for model development. We reported the values of macro AUC-ROC with 95\% confidence intervals on the testing sets, TCGA and TTH.}
\label{tab4}
\end{table}
Among the two additional modalities, we found that removing texts decreased the performance the most, especially on tissue type and procedure type metadata prediction (Table~\ref{tab4}, $\Delta$(All w/o text, All) = -14.94\% (tissue), -2.28\% (fixation), -2.16\% (procedure), -1.14\% (staining)).
Case-level structured data is also predictive for some tasks but not as informative as texts (Table~\ref{tab4}, $\Delta$(All w/o structured, All) = -4.79\% (tissue), -2.84\% (fixation), -5.93\% (procedure), -1.17\% (staining)).
The observed trend is consistent with the understanding that the pathology reports contain information closely related to diagnosis and tissue type but not fixation, staining information.

\paragraph{Additional Informative Modality Might Not Be Helpful}
For research on multimodal modeling, a common understanding is that adding informative modality is helpful for prediction. 
However, we observed inferior performance after adding an informative modality. 
Specifically, we explored incorporating high magnification image patches to improve the performance across all four tasks. 
With only image modality as the input, patch information helps the model perform better in three out of four tasks (Table~\ref{tab5}, $\Delta$(Image, Image w/o patch) = 0.86\% (tissue), 5.15\% (fixation), -2.26\% (procedure), 0.6\% (staining)).
However, adding patches yields inferior performance for the tissue and procedure type prediction when text and structured data are also used (Table~\ref{tab5}, $\Delta$(All, All w/o patch) = -10.77\% (tissue), 2.27\% (fixation), -11.69\% (procedure), 1.16\% (staining)).
\begin{table}[]
\centering
\resizebox{\linewidth}{!}
{
\begin{tabular}{cccccc}
\toprule
Dataset & Modality & Tissue & Fixation & Procedure & Staining \\
\midrule
TCGA  & Image & \textbf{0.58 (0.56, 0.61)} & \textbf{0.86 (0.84, 0.88)} & 0.67 (0.62, 0.73) & 0.86 (0.15, 1.00) \\
 (external)  & Image w/o patch & 0.57 (0.55, 0.59) & 0.78 (0.75, 0.81) & \textbf{0.72 (0.67, 0.77)} & \textbf{0.88 (0.40, 1.00)} \\ \cline{2-6} 
 & All & 0.81 (0.79, 0.83) & \textbf{0.88 (0.85, 0.90)} & 0.67 (0.61, 0.72) & \textbf{0.88 (0.24, 1.00)} \\
 & All w/o patch & \textbf{0.92 (0.91, 0.93)} & 0.84 (0.82, 0.86) & \textbf{0.77 (0.72, 0.81)} & 0.87 (0.37, 1.00) \\
\midrule
TTH & Image & \textbf{0.75 (0.69, 0.80)} & \textbf{0.99 (0.98, 1.00)} & \textbf{0.68 (0.60, 0.76)} & \textbf{0.85 (0.78, 0.91)} \\
 & Image w/o patch & 0.71 (0.66, 0.76) & 0.98 (0.94, 1.00) & 0.66 (0.57, 0.74) & 0.82 (0.73, 0.89) \\ \cline{2-6} 
 & All & 0.88 (0.85, 0.91) & 0.98 (0.94, 1.00) & 0.71 (0.63, 0.79) & \textbf{0.84 (0.77, 0.90)} \\
 & All w/o patch & \textbf{0.95 (0.92, 0.96)} & \textbf{0.98 (0.95, 1.00)} & \textbf{0.77 (0.69, 0.84)} & 0.83 (0.76, 0.90) \\ 
\bottomrule
\end{tabular}
}
\caption{Ablation analysis on multiscale image information on the testing set, TCGA and TTH. We reported the values of macro AUC-ROC with 95\% confidence intervals.}
\label{tab5}
\end{table}

We argue that the inferior performance is due to the similarity in information between patch images and texts/structured data. 
Patch image is expected to be noisier than the other two modalities primarily due to image noise and patch sampling noise.
The patch image only contains a narrow specific region. Therefore, the noise might come from capturing tissue images with similar visual features that are common across different tissue types. 
Since we only picked up three random patches, the patches may not be representative enough for a specific tissue type.
This finding also leads to a potential future direction of developing a method to identify the label-specific region for patch generation without extensive annotation. 
Please refer to Appendix A5 for detailed discussions from a pathologist's perspective.


\section{Conclusions}
Due to the need and challenges of acquiring well-curated metadata for large scale biobanking, in this study, we explored the potential for using a machine learning approach for slide-level metadata prediction.
We proposed a multimodal multitask model to leverage information across different modalities for the prediction of several important metadata jointly. 
The results showed that the proposed framework outperformed single modality single task baseline. 
It also showed better performance when generalizing to the independent TCGA test set from a different source.
This generalization is particularly important because its a better estimation of the true performance on other future unseen datasets. 
We expect this model to be useful for increasing the utilization of existing biobank archives, and the proposed framework to be a helpful reference for future multimodal multitask learning research in the biomedical space.

\section{Acknowledgment}
The authors thank Dr. Tiam Jaroensri for detailed comments and reviews, and all the members in Google Health for their kind suggestions.
The results published are in whole or part based upon data generated by the TCGA Research Network: \texttt{https://www.cancer.gov/tcga}.

\clearpage
\bibliography{mybib}
\bibliographystyle{plain}

\clearpage
\setcounter{table}{0}
\setcounter{subsection}{0}
\section*{Appendix}

\paragraph{A1 Usage of Case-level Structured Data}
\label{a1}
We used the primary cancer site of the patient as the source of case-level structured data. 
Even though the primary cancer site seems to be similar to the tissue type, they are different from the pathology perspective.
Primary cancer site is the case-level information yet tissue type is slide-level information (Figure~\ref{fig1}, Appendix Table A1).
The tissue type of a slide can be completely different from the case-level primary cancer site even though they are from the same patient.
For example, a patient with the primary cancer site of the breast may have multiple pathology slides.
Some of the slides can be lymph node or skin tissue due to metastasis and invasion. 
\begin{table}[h]
\resizebox{\columnwidth}{!}{
\begin{tabular}{cccc}
\toprule
 & Visit & Generated data &  \\
\midrule
Time & Reason & Case-level & Slide-level \\
\midrule
1 & \begin{tabular}[c]{@{}c@{}}Breast mass, \\ arrange biopsy\end{tabular} & Clinical note$\times$1 &  \\
\midrule
2 & Biopsy & \begin{tabular}[c]{@{}c@{}}Pathology report$\times$1\\ (benign)\end{tabular} & Slide images$\times$2 \\
\midrule
3 & Followup & Clinical note$\times$1 &  \\
\midrule
4 & \begin{tabular}[c]{@{}c@{}}Another mass, \\ arrange biopsy\end{tabular} & Clinical note$\times$1 &  \\
\midrule
5 & Biopsy & \begin{tabular}[c]{@{}c@{}}Pathology report$\times$1\\ (high grade IDC)\end{tabular} & Slide images$\times$3 \\
\midrule
6 & \begin{tabular}[c]{@{}c@{}}Followup, \\ arrange surgery\end{tabular} & Clinical note$\times$1 &  \\
\midrule
7 & Surgery & \begin{tabular}[c]{@{}c@{}}Pathology report$\times$1\\ (IDC, stage 2a, T2N1M0)\\ surgery report$\times$1\end{tabular} & \begin{tabular}[c]{@{}c@{}}Slide images$\times$10 \\ (both breast and lymph node tissues, \\ with frozen and FFPE fixations, \\ H\&E and IHC staining)\end{tabular} \\
\bottomrule
\end{tabular}
}
\caption{An example of patient visits from complaining a breast mass to surgery. The pathology data in case-level and slide-level provide different information of the disease.}
\label{stab1}
\end{table}

\paragraph{A2 Derivation of the Multi-objective Loss}
\label{a2}
Given the metadata prediction task $t$, we make the assumption that the task predicted output $y_{it}$ for the training input $x_i$ returned by the neural network $f_t$ under the weight set $w$, the noise scalar (error) of the output is zero-mean and normally distributed with the standard deviation of $\sigma_t$. 
Thus, we want to find the stationary points of the Gaussian log likelihood:
\begin{equation}
\sum_{it} \log(\frac{1}{\sqrt{2\pi} \sigma_t} \exp(-\frac{(y_{it} - f_t(x_i; w))^2}{2 \sigma_t^2}))
\end{equation}

We set the partial derivation of the Gaussian log likelihood with respect to the variance $\sigma^2$ to zero and obtain the update equation for $\sigma^2$:
\begin{equation}
\frac{\partial}{\partial \sigma_t^2} \sum_i \log(\frac{1}{\sqrt{2\pi} \sigma_t} \exp(-\frac{(y_{it} - f_t(x_i; w))^2}{2 \sigma_t^2})) = 0
\end{equation}

\begin{equation}
\frac{\partial}{\partial \sigma_t^2} \sum_i (-\frac{\log \sigma_t^2}{2} - \frac{(y_{it} - f_t(x_i; w))^2}{2 \sigma_t^2}) = 0
\end{equation}

\begin{equation}
\sum_i (-\frac{1}{2\sigma_t^2} + \frac{(y_{it} - f_t(x_i; w))^2}{2 \sigma_t^4}) = 0
\end{equation}

\begin{equation}
\sum_i (-1 + \frac{(y_{it} - f_t(x_i; w))^2}{\sigma_t^2}) = 0
\end{equation}

\begin{equation}
\sigma_t^2 = \frac{\sum_i (y_{it} - f_t(x_i; w))^2}{N},
\end{equation}
where $N$ is the size of training data. Then we replace the$\sigma_t^2$ in the log likelihood function and get the following simplified version:

\begin{equation}
\sum_t (-\frac{N}{2} \log (\frac{1}{N} \sum_i (y_{it} - f_t(x_i; w))^2) - \frac{N}{2})
\end{equation}

\begin{equation}
- \sum_t \log (\sum_i (y_{it} - f_t(x_i; w))^2).
\end{equation}

To maximize the likelihood, we convert the above to the format of loss function and minimize it:
\begin{equation}
\mathcal{L}_\mathrm{multi} = \sum_t \log ( \sum_i (y_{it} - f_t(x_i; w))^2),
\end{equation}

The above loss function can be optionally exponentiate back as:
\begin{equation}
\mathcal{L}_\mathrm{multi} = \prod_t \sum_i (y_{it} - f_t(x_i; w))^2
\end{equation}

Where we can see the loss is a geometric average instead of the arithmetic mean.

\begin{table}[h]
\resizebox{\columnwidth}{!}{
\begin{tabular}{cl}
\toprule
Abbreviation & Description \\
\midrule
AUC-ROC & Area under receiver operating characteristic curve \\
AUC-PR & Area under precision-recall curve \\
BERT & Bidirectional Encoder Representations from Transformers \\
CBP & Compact bilinear pooling \\
CE & Cross entropy \\
CSP & Count sketch projection \\
FFPE & Formalin-fixed paraffin-embedded fixation \\
FFT & Fourier transformation \\
H\&E & Hematoxylin and eosin staining \\
H\&N & Head and neck tissue \\
IHC & Immunohistochemical staining \\
LN & Lymph node tissue \\
MTL & Multitask learning \\
TCGA & The Cancer Genome Atlas \\
TTH & Tertiary teaching hospital dataset \\
UGI & Upper gastrointestinal tissue \\
\bottomrule
\end{tabular}
}
\caption{Dictionary of abbreviations.}
\label{stab2}
\end{table}

\paragraph{A3 Evaluation of Model Performance}
\label{a3}
Except for the model performance demonstrated in the main context (Table 2-5), we provide the additional model evaluation results in this supplemental section. 
For different experiments in the main context, we reported the values of AUC-PR, which consider the trade-off between precision and recall, on both TCGA and TTH test sets in Table A3, A5, A7, A9.
We also demonstrate the model evaluation of both AUC-ROC and AUC-PR on the validation set in Table A4, A6, A8, A10.

\begin{table}[]
\centering
\resizebox{\linewidth}{!}
{
\begin{tabular}{cccccc}
\toprule
Dataset & Cfg & Tissue & Fixation & Procedure & Staining \\
\midrule
TCGA & SS & 0.07 (0.05, 0.10) & 0.85 (0.82, 0.87) & 0.57 (0.54, 0.60) & 0.74 (0.50, 1.00) \\
 & SM & 0.15 (0.13, 0.17) & 0.83 (0.80, 0.85) & 0.56 (0.53, 0.60) & 0.73 (0.50, 1.00) \\
 & MS & 0.21 (0.15, 0.29) & 0.91 (0.89, 0.93) & 0.56 (0.53, 0.60) & 0.66 (0.50, 1.00) \\
 & MM & 0.37 (0.34, 0.40) & 0.87 (0.84, 0.89) & 0.57 (0.54, 0.60) & 0.72 (0.50, 1.00) \\ 
\midrule
TTH & SS & 0.36 (0.21, 0.52) & 0.97 (0.87, 1.00) & 0.67 (0.59, 0.74) & 0.82 (0.74, 0.89) \\
 & SM & 0.41 (0.32, 0.50) & 0.97 (0.88, 1.00) & 0.66 (0.57, 0.74) & 0.83 (0.75, 0.90) \\
 & MS & 0.63 (0.47, 0.77) & 0.87 (0.73, 0.99) & 0.67 (0.59, 0.76) & 0.80 (0.73, 0.87) \\
 & MM & 0.59 (0.51, 0.66) & 0.94 (0.84, 1.00) & 0.71 (0.63, 0.78) & 0.81 (0.73, 0.88) \\ 
\bottomrule
\end{tabular}
}
\caption{Performance between different modality-task settings on the testing set, TCGA and TTH. TCGA is specifically for external validation, which has the different data distribution from the training set. We reported the values of macro AUC-PR with 95\% confidence intervals. SS: single modal single task; SM: single modal multitask; MS: multimodal single task; MM: multimodal multitask.}
\label{stab3}
\end{table}

\begin{table}[]
\centering
\resizebox{\linewidth}{!}
{
\begin{tabular}{cccccc}
\toprule
Metric & Cfg & Tissue & Fixation & Procedure & Staining \\
\midrule
AUC-ROC & SS & 0.72 (0.61, 0.82) & 0.99 (0.94, 1.00) & 0.70 (0.61, 0.78) & 0.83 (0.74, 0.90) \\
 & SM & 0.78 (0.72, 0.82) & 0.99 (0.95, 1.00) & 0.69 (0.60, 0.78) & 0.83 (0.75, 0.90) \\
 & MS & 0.90 (0.85, 0.95) & 0.95 (0.83, 1.00) & 0.70 (0.62, 0.78) & 0.81 (0.72, 0.89) \\
 & MM & 0.90 (0.86, 0.92) & 0.98 (0.91, 1.00) & 0.73 (0.63, 0.81) & 0.82 (0.74, 0.89) \\ 
\midrule
AUC-PR & SS & 0.35 (0.20, 0.50) & 0.96 (0.85, 1.00) & 0.70 (0.61, 0.79) & 0.79 (0.71, 0.87) \\
 & SM & 0.45 (0.35, 0.53) & 0.97 (0.89, 1.00) & 0.69 (0.60, 0.77) & 0.80 (0.72, 0.88) \\
 & MS & 0.63 (0.46, 0.78) & 0.87 (0.70, 0.98) & 0.70 (0.62, 0.78) & 0.78 (0.70, 0.86) \\
 & MM & 0.62 (0.53, 0.70) & 0.93 (0.81, 1.00) & 0.72 (0.64, 0.81) & 0.79 (0.70, 0.86) \\ 
\bottomrule
\end{tabular}
}
\caption{Performance between different modality-task settings on the validation set. We reported both the values of macro AUC-ROC and AUC-PR with 95\% confidence intervals. SS: single modal single task; SM: single modal multitask; MS: multimodal single task; MM: multimodal multitask.}
\label{stab4}
\end{table}

\begin{table}[]
\centering
\resizebox{\linewidth}{!}
{
\begin{tabular}{cccccc}
\toprule
Dataset & MM-Strategy & Tissue & Fixation & Procedure & Staining \\
\midrule
TCGA & Concat & 0.37 (0.34, 0.40) & 0.87 (0.84, 0.89) & 0.57 (0.54, 0.60) & 0.72 (0.50, 1.00) \\
 & CBP & 0.74 (0.73, 0.76) & 0.52 (0.49, 0.55) & 0.52 (0.51, 0.54) & 0.63 (0.50, 1.00) \\
\midrule
TTH & Concat & 0.59 (0.51, 0.66) & 0.94 (0.84, 1.00) & 0.71 (0.63, 0.78) & 0.81 (0.73, 0.88) \\
 & CBP & 0.64 (0.57, 0.70) & 0.52 (0.48, 0.57) & 0.53 (0.47, 0.61) & 0.49 (0.44, 0.55) \\
\bottomrule
\end{tabular}
}
\caption{Performance on the testing set, TCGA and TTH. TCGA is specifically for external validation, which has the different data distribution from the training set. We reported the values of macro AUC-PR with 95\% confidence intervals.}
\label{stab5}
\end{table}

\begin{table}[]
\centering
\resizebox{\linewidth}{!}
{
\begin{tabular}{cccccc}
\toprule
Metric & MT-Strategy & Tissue & Fixation & Procedure & Staining \\
\midrule
AUC-ROC & Concat & 0.90 (0.86, 0.92) & 0.98 (0.91, 1.00) & 0.73 (0.63, 0.81) & 0.82 (0.74, 0.89) \\
 & CBP & 0.90 (0.87, 0.93) & 0.51 (0.29, 0.74) & 0.52 (0.43, 0.62) & 0.47 (0.38, 0.58) \\
\midrule
AUC-PR & Concat & 0.62 (0.53, 0.70) & 0.93 (0.81, 1.00) & 0.72 (0.64, 0.81) & 0.79 (0.70, 0.86) \\
 & CBP & 0.66 (0.57, 0.72) & 0.52 (0.48, 0.58) & 0.55 (0.48, 0.63) & 0.50 (0.45, 0.57) \\
\bottomrule
\end{tabular}
}
\caption{Performance on the validation set. We reported both the values of macro AUC-ROC and AUC-PR with 95\% confidence intervals.}
\label{stab6}
\end{table}

\begin{table}[]
\centering
\resizebox{\linewidth}{!}
{
\begin{tabular}{cccccc}
\toprule
Dataset & Modality & Tissue & Fixation & Procedure & Staining \\
\midrule
TCGA & All & 0.37 (0.34, 0.40) & 0.87 (0.84, 0.89) & 0.57 (0.54, 0.60) & 0.72 (0.50, 1.00) \\
 & All w/o text & 0.20 (0.17, 0.22) & 0.83 (0.81, 0.86) & 0.57 (0.54, 0.60) & 0.71 (0.50, 1.00) \\
 & All w/o structured & 0.29 (0.26, 0.31) & 0.85 (0.82, 0.87) & 0.54 (0.52, 0.58) & 0.70 (0.50, 1.00) \\
\midrule
TTH & All & 0.59 (0.51, 0.66) & 0.94 (0.84, 1.00) & 0.71 (0.63, 0.78) & 0.81 (0.73, 0.88) \\
 & All w/o text & 0.42 (0.34, 0.50) & 0.93 (0.83, 1.00) & 0.67 (0.59, 0.76) & 0.81 (0.74, 0.88) \\
 & All w/o structured & 0.54 (0.46, 0.62) & 0.93 (0.81, 1.00) & 0.69 (0.60, 0.76) & 0.81 (0.72, 0.88) \\
\bottomrule
\end{tabular}
}
\caption{Performance on the testing set, TCGA and TTH. TCGA is specifically for external validation, which has the different data distribution from the training set. We reported the values of macro AUC-PR with 95\% confidence intervals.}
\label{stab7}
\end{table}

\begin{table}[]
\centering
\resizebox{\linewidth}{!}
{
\begin{tabular}{cccccc}
\toprule
Metric & Modality & Tissue & Fixation & Procedure & Staining \\
\midrule
AUC-ROC & All & 0.90 (0.86, 0.92) & 0.98 (0.91, 1.00) & 0.73 (0.63, 0.81) & 0.82 (0.74, 0.89) \\
 & All w/o text & 0.81 (0.76, 0.85) & 0.98 (0.92, 1.00) & 0.69 (0.61, 0.77) & 0.82 (0.74, 0.89) \\
 & All w/o structured & 0.87 (0.83, 0.90) & 0.98 (0.94, 1.00) & 0.70 (0.63, 0.78) & 0.81 (0.74, 0.88) \\
\midrule
AUC-PR & All & 0.62 (0.53, 0.70) & 0.93 (0.81, 1.00) & 0.72 (0.64, 0.81) & 0.79 (0.70, 0.86) \\
 & All w/o text & 0.46 (0.36, 0.55) & 0.92 (0.80, 1.00) & 0.69 (0.61, 0.77) & 0.78 (0.70, 0.86) \\
 & All w/o structured & 0.56 (0.46, 0.64) & 0.93 (0.82, 1.00) & 0.70 (0.62, 0.78) & 0.78 (0.69, 0.85) \\
\bottomrule
\end{tabular}
}
\caption{Performance on the validation set. We reported both the values of macro AUC-ROC and AUC-PR with 95\% confidence intervals.}
\label{stab8}
\end{table}

\begin{table}[]
\centering
\resizebox{\linewidth}{!}
{
\begin{tabular}{cccccc}
\toprule
Dataset & Modality & Tissue & Fixation & Procedure & Staining \\
\midrule
TCGA & Image & 0.14 (0.13, 0.16) & 0.85 (0.82, 0.87) & 0.57 (0.54, 0.61) & 0.74 (0.50, 1.00) \\
 & Image w/o patch & 0.12 (0.11, 0.13) & 0.77 (0.73, 0.80) & 0.59 (0.56, 0.63) & 0.67 (0.50, 1.00) \\ \cline{2-6}
 & All & 0.37 (0.34, 0.40) & 0.87 (0.84, 0.89) & 0.57 (0.54, 0.60) & 0.72 (0.50, 1.00) \\
 & All w/o patch & 0.64 (0.61, 0.66) & 0.84 (0.81, 0.86) & 0.61 (0.57, 0.65) & 0.68 (0.50, 1.00) \\
\midrule
TTH & Image & 0.40 (0.31, 0.48) & 0.97 (0.89, 1.00) & 0.67 (0.59, 0.75) & 0.82 (0.75, 0.90) \\
 & Image w/o patch & 0.29 (0.22, 0.37) & 0.93 (0.81, 1.00) & 0.65 (0.57, 0.73) & 0.79 (0.70, 0.87) \\ \cline{2-6}
 & All & 0.59 (0.51, 0.66) & 0.94 (0.84, 1.00) & 0.71 (0.63, 0.78) & 0.81 (0.73, 0.88) \\
 & All w/o patch & 0.74 (0.67, 0.81) & 0.91 (0.79, 0.99) & 0.76 (0.69, 0.83) & 0.81 (0.74, 0.89) \\ 
\bottomrule
\end{tabular}
}
\caption{Performance on the testing set, TCGA and TTH. TCGA is specifically for external validation, which has the different data distribution from the training set. We reported the values of macro AUC-PR with 95\% confidence intervals.}
\label{stab9}
\end{table}

\begin{table}[]
\centering
\resizebox{\linewidth}{!}
{
\begin{tabular}{cccccc}
\toprule
Metric & Modality & Tissue & Fixation & Procedure & Staining \\
\midrule
AUC-ROC & Image & 0.78 (0.73, 0.83) & 0.99 (0.94, 1.00) & 0.70 (0.61, 0.78) & 0.83 (0.75, 0.90) \\
 & Image w/o patch & 0.74 (0.69, 0.79) & 0.98 (0.91, 1.00) & 0.69 (0.60, 0.77) & 0.78 (0.70, 0.86) \\ \cline{2-6} 
 & All & 0.90 (0.86, 0.92) & 0.98 (0.91, 1.00) & 0.73 (0.63, 0.81) & 0.82 (0.74, 0.89) \\
 & All w/o patch & 0.95 (0.93, 0.97) & 0.98 (0.95, 1.00) & 0.79 (0.71, 0.85) & 0.81 (0.72, 0.88) \\ 
\midrule
AUC-PR & Image & 0.44 (0.35, 0.53) & 0.96 (0.86, 1.00) & 0.69 (0.61, 0.78) & 0.80 (0.70, 0.87) \\
 & Image w/o patch & 0.31 (0.23, 0.40) & 0.91 (0.76, 1.00) & 0.68 (0.60, 0.77) & 0.76 (0.66, 0.85) \\ \cline{2-6} 
 & All & 0.62 (0.53, 0.70) & 0.93 (0.81, 1.00) & 0.72 (0.64, 0.81) & 0.79 (0.70, 0.86) \\
 & All w/o patch & 0.77 (0.70, 0.83) & 0.90 (0.71, 0.99) & 0.78 (0.70, 0.85) & 0.79 (0.70, 0.87) \\ 
\bottomrule
\end{tabular}
}
\caption{Performance on the validation set. We reported both the values of macro AUC-ROC and AUC-PR with 95\% confidence intervals.}
\label{stab10}
\end{table}

\begin{figure*}[htbp]
\centering
\includegraphics[width=1.0\linewidth]{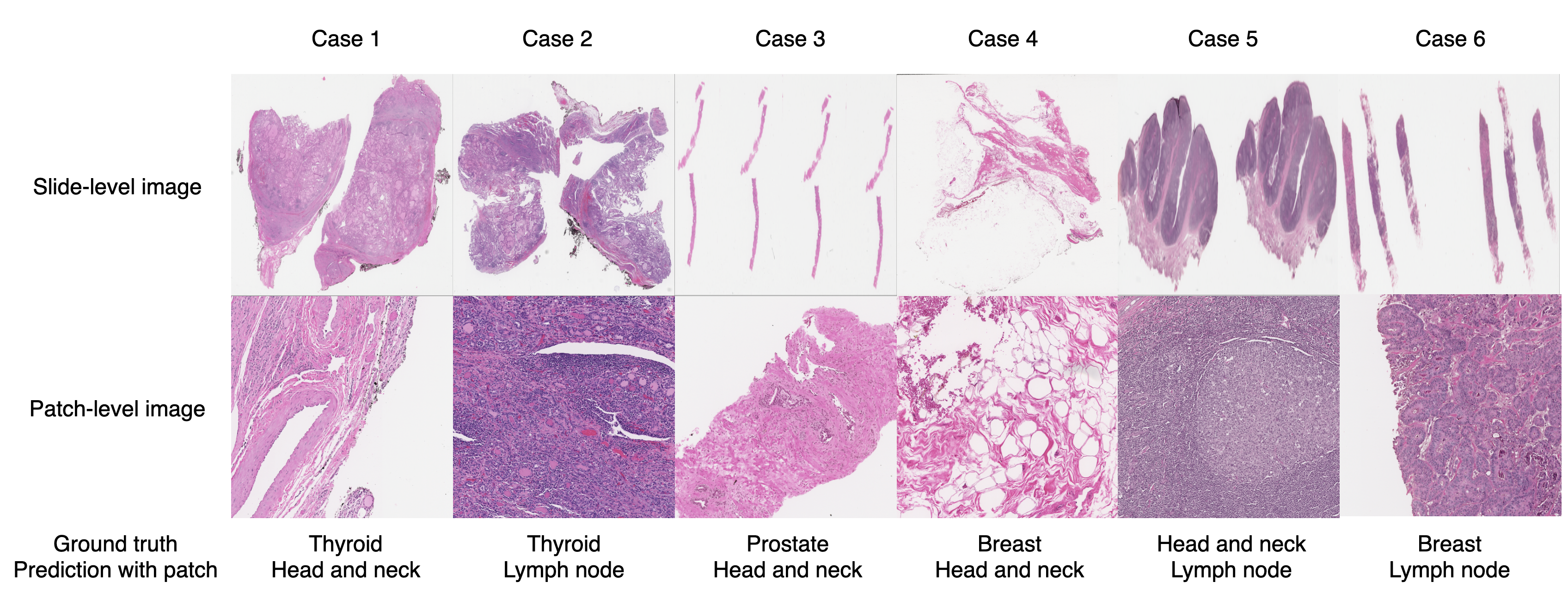} 
\hfill
\caption{Examples of the cases that the models with patch integration fail.}
\label{sfig1}
\end{figure*}

\paragraph{A4 Relations between Multiscale Imaging and Prediction Tasks}
\label{a4}
Both slide- and patch-level images are essential for the metadata prediction from the pathology perspective.

The tissue type prediction requires patch images to observe cellular and regional tissue information. 
This is hard at the whole slide level at low magnification. 
Since the same tissue type specimen can be processed in various ways, it can eventually result in different shapes at low magnification and therefore the whole slide image can't be a good pattern for identifying the tissue type.
High magnification images that demonstrate cell morphology and tissue structure are essential if there are no additional modalities used.

For the fixation type prediction, it is difficult to identify it at whole slide level but relatively easier in the patch since the intracellular matrix in frozen section is usually not preserved well and fragmented at high magnification due to the fast but less delicate tissue fixation process. 
Also, the frozen section staining process is not robust as the FFPE fixation and therefore the images usually have less contrast. 
However, the contrast issue can be found not only because of the fixation type but also other issues, such as stain normalization problem. 
Therefore, patch-level images are still required for better fixation type prediction.

Identifying the procedure type is challenging at both slide and patch level due to the definition ambiguity of the procedure. 
The biopsy can be needle core biopsy, incisional biopsy or excisional biopsy. 
The latter two biopsy types can be very similar to surgical resection, and therefore hard to be discriminated from each other. 
This is also challenging for an annotation since every pathologist and specialist has a different interpretation of the procedure. 
For example, we usually use incisional/excisional biopsy for skin surgery rather than calling it major resection.
Instead of image modality, we may rely on other inputs such as free text reports to improve model performance.

Staining type prediction has fewer issues but it highly depends on color normalization across hospitals and labs.
Even for the same hospitals, preservation and timing are also critical for the staining quality.

\paragraph{A5 Pathology Perspective of Usage of Multiscale Imaging}
\label{a5}
In Table~\ref{tab5}, we identified that patch integration didn't work well while other modalities (text and structured data) were used.
We consulted the board-certified pathologists to ensure our findings and interpretation are reasonable.
We demonstrate some examples in Figure~\ref{sfig1} that were misclassified by the model using patch information, yet correctly classified by the model without the patch.

As we mentioned in the main context, the patch information may not be representative enough through the generation process, which is commonly used for most machine learning tasks in pathology.
For example, we expect to identify the breast tissue by seeing the breast epithelial or tubular structures. 
However, the patch may focus solely on fat, muscular or connective tissues (e.g. collagen fibers), which are general across many tissue types and therefore tend to be misclassified to the head and neck tissue, which has all these features in the submucosal layer (Figure~\ref{sfig1} Case 1, 3, 4).

The case with the patches that are cell-abundant may also tend to be misclassified. 
For instance, the dense regions of the thyroid tissue are similar to those in the lymph node tissue (Figure~\ref{sfig1} Case 2). 
Even more, there are some regions in head and neck is gathered the lymphatic cells (Figure~\ref{sfig1} Case 5).
Including such correct but misleading patches may bias the result toward the incorrect prediction. i.e. from head and neck tissue to lymph node tissue.

Finally, the patch of cancer metastasis is also a source of misclassification between primary cancer organ and metastatic site. e.g. breast and lymph node (Figure~\ref{sfig1} Case 6).
Even though the patch information is helpful while other modalities such as reports are not used, further patch processing is required to obtain many representative patches for better integration.

\end{document}